\documentclass[journal,onecolumn]{IEEEtran}
\usepackage{times}  
\usepackage{helvet} 
\usepackage{courier}  
\usepackage[hyphens]{url}  
\usepackage{graphicx} 
\usepackage{subcaption}     
\usepackage{mathrsfs,bbm,bm}
\usepackage{microtype}      
\usepackage[utf8]{inputenc} 
\usepackage{url}            
\usepackage{booktabs}       
\usepackage{amsfonts}       
\usepackage{nicefrac}       
\usepackage{amsmath}        

\ifCLASSINFOpdf
\else
\fi

\begin{document}

%
\title{Learning Various Length Dependence by Dual Recurrent Neural Networks}
%
%
%

\author{Chenpeng~Zhang,~
        Shuai~Li,~\IEEEmembership{Member,~IEEE},
        ~Mao~Ye$^*$, ~\IEEEmembership{Member,~IEEE}, 
        ~Ce~Zhu, ~\IEEEmembership{Fellow,~IEEE},
        and ~Xue~Li, ~\IEEEmembership{Member,~IEEE}
\thanks{Chenpeng Zhang and Mao Ye are with the School of Computer Science and Engineering, University of Electronic Science and Technology of China, Chengdu 611731, P.R. China (e-mail: cvlab.uestc@gmail.com).}
\thanks{Shuai Li and Ce Zhu are with the School of Information and Communication Engineering, University of Electronic Science and Technology of China, Chengdu 611731, P.R. China.}
\thanks{Xue Li is with the School of Information Technology and Electronic Engineering, The University of Queensland, Brisbane, QLD 4072, Australia.}
\thanks{*corresponding author}
}

\maketitle

\begin{abstract}
Recurrent neural networks (RNNs) are widely used as a memory model for sequence-related problems.
Many variants of RNN have been proposed to solve the gradient problems of training RNNs and process long sequences.
Although some classical models have been proposed, capturing long-term dependence while responding to short-term changes remains a challenge.
To this problem, we propose a new model named Dual Recurrent Neural Networks (DuRNN).
The DuRNN consists of two parts to learn the short-term dependence and progressively learn the long-term dependence.
The first part is a recurrent neural network with constrained full recurrent connections to deal with short-term dependence in sequence and generate short-term memory.
Another part is a recurrent neural network with independent recurrent connections which helps to learn long-term dependence and generate long-term memory.
A selection mechanism is added between two parts to help the needed long-term information transfer to the independent neurons.
Multiple modules can be stacked to form a multi-layer model for better performance.
Our contributions are: 1) a new recurrent model developed based on the divide-and-conquer strategy to learn long and short-term dependence separately, and 2) a selection mechanism to enhance the separating and learning of different temporal scales of dependence.
Both theoretical analysis and extensive experiments are conducted to validate the performance of our model, and we also conduct simple visualization experiments and ablation analyses for the model interpretability.
Experimental results indicate that the proposed DuRNN model can handle not only very long sequences (over 5000 time steps), but also short sequences very well.
Compared with many state-of-the-art RNN models, our model has demonstrated efficient and better performance.
\end{abstract}


%
\IEEEpeerreviewmaketitle

\section{Introduction}

\IEEEPARstart{R}{ecurrent} neural networks (RNNs) are usually used for processing sequence-related problems as a memory model.
They widely appear in deep learning models to solve problems such as sequence learning \cite{sutskever2014sequence}, language modeling \cite{cho2014learning}, etc.
These neural networks are designed to approach a dynamical system to get a complex time relationship, which can also be called as sequential dependence.
A RNN structure has the connections between hidden units which can be described as the following \cite{pascanu2013difficulty},
\begin{equation*}
	\bm{h}_{t}=\sigma(\bm{W}_{in}\bm{x}_{t}+\bm{W}_{rec}\bm{h}_{t-1}+\bm{b}),
\end{equation*}
where $\bm{x}_{t}\in\mathbb{R}^M$ and $\bm{h}_{t}\in\mathbb{R}^N$, are the input and the hidden state at time $t$, respectively.
$\bm{W}_{in}\in\mathbb{R}^{N\times{M}}$, $\bm{W}_{rec}\in\mathbb{R}^{N\times{N}}$, and $\bm{b}\in\mathbb{R}^N$ are the variables to be trained  which are the input weight matrix, the recurrent weight matrix and the bias, respectively.
$\sigma$ is the element-wise activation function.
$M$ and $N$ are the dimension of the input feature and the number of hidden neurons, respectively.

RNNs cannot process long-term sequences because of the gradient vanishing and exploding problems due to the improper activation function and the uncontrolled repeated multiplications of the recurrent weight matrix \cite{pascanu2013difficulty}.
The existing methods proposed to solve the problem of learning long-term dependence can be roughly divided into two categories: the gated RNNs and the gate-free RNNs.
The first category of methods enhance the original recurrent neurons with a set of gates to preserve long-term memory,
for example, the famous long short-term memory (LSTM) and GRU models
\cite{hochreiter1997long,Cho2014On}.  Although the gated RNNs get notable performance and have become the popular configuration for many machine learning tasks, the generation of gates takes up considerable amount of computing resources and the forget gate also limits the learning of very long sequences while the uncontrolled recurrent weight matrix can cause gradient exploding \cite{kanai2017preventing}. The second category of methods are gate-free RNNs proposed to solve the gradient problems without using gates. While the gate-free RNNs perform better on learning the long-term dependence, less efforts have been devoted to reducing the short-term disturbances.

In the study \cite{chung2017hierarchical}, long-term dependence is expected to be learned by the deep layers of a multi-layer HMLSTM network while the shallow layers of it are used for the short-term dependence.
This rule is usually utilized unconsciously when constructing multi-layer RNN model to learn multi-scale dependence.
However, simply using the same kind of LSTM cells is hard to adapt to different functional requirements.
On the contrary, using different models for long-term and short-term dependence may be more adaptive.
Based on this motivation, a Dual Recurrent Neural Networks (DuRNN) is proposed in this paper.
Consisting of two parts and a transition connection with selection mechanism between them, DuRNN progressively learns the short and long-term dependence. The first part is a conventional recurrent neural network with full connected neurons, where the singular values of the recurrent weight matrix are constrained to learn short-term memory and avoid gradient exploding. The second part consists of independent recurrent connected neurons \cite{li2018independently} with numerical restriction on the recurrent weights to learn long-term dependence. The non-saturated activation function relu \cite{M2013On} is used in both parts.
A selection mechanism is added to transfer the different dependence learned by the two parts in order to complete the divide-and-conquer strategy of learning separate dependence.
Similar to the macroscopic human memory generative processes, i.e., humans first have the short-term memory and then transfer it to the long-term memory \cite{Multi}, we are trying to adapt a divide-and-conquer strategy to learn long-term information.
Similar to the state-of-the-art CNN and RNN methods for generating deep models, our dual recurrent neural networks can also be considered as a layer to be stacked to form a multi-layer DuRNN for better performance.

Our contributions can be summarized as follows.

1) A new framework is proposed to assist the capture of long-term dependence.
It uses two kinds of recurrent connections, i.e, local full recurrent connections and long-term independent recurrent connections to progressively learn the short-term and long-term dependence.

2) We propose a selection mechanism to connect these different recurrent networks such that the long-term memory can be better separated and utilized.

3) Extensive experiments are conducted on the adding problem,  MNIST classification, language modeling and action recognition. The results confirm that DuRNN improves the performance in both long and short sequences.

\section{Related Works}
\subsection{Gated Recurrent Neural Networks}

LSTM models are popular for temporal information processing \cite{alom2019a}.
After the long short-term memory (LSTM) \cite{hochreiter1997long} model was proposed, many variations of LSTM have been developed. Roughly, there exist three approaches, i.e., improvements around the gated units, cell neurons or additional mechanism based.

The first category of methods try to reduce the number of gates and trainable variables such as GRU \cite{Cho2014On} and MRU \cite{zhou2016minimal}.
With the modified gates, this kind of methods are easier to achieve better performance on a variety of tasks.
The second category improves around the cell state updating, such as ON-LSTM \cite{shen2019ordered}.
The last category of methods use some additive mechanisms such that special dependence or non-sequential information can be learned.
For example, the model STA-LSTM  pays different levels of attention to the dependence of discriminative joints of skeleton \cite{Song2017An}; convolutional LSTM \cite{shi2015convolutional} encodes spatial information while processing sequences, etc.
Furthermore, techniques such as recurrent dropout, recurrent batch normalization, zoneout are always used to improve the performance of gated RNNs models \cite{Semeniuta2016Recurrent,Cooijmans2017Recurrent,Krueger2017Zoneout}, and works like \cite{lobacheva2020structured} and \cite{Tanaka2020Spatially} which sparse the gated models also provide a performance improvement.

Though massive improvements have already been made, these approaches are still relatively complex and sequential dependence is still difficult to be captured.
At the same time, the gradient back propagation of very long sequence is also not guaranteed to be stable without vanishing because the uncontrolled forget gates may cause it to decrease.
So these approaches cannot process very long sequences well while learning the sequence with general length is rather effective.

\subsection{Gate-free Recurrent Neural Networks}
On the contrary, gate-free RNNs put almost all their attentions on long-term dependence learning.

Some models restrict the recurrent weight matrix and the activation function to capture the long-term dependence. For example, IRNN and uRNN use relu as their activation functions \cite{le2015simple,arjovsky2016unitary}.
IRNN has a positive definite and identical initialization for recurrent weight matrix such that gradient vanishing will be mitigated; and the recurrent weight matrix of uRNN is required to be orthogonal to totally control the gradient propagations.
Some approaches try to establish cross-scale connection to make the long-term dependence as easy to being learned as the short one \cite{Koutn2014A}.
Some methods like RNN-path \cite{Neyshabur2016Path} develop a path adaptation of SGD optimization method to avoid the gradient problems caused by BPTT \cite{Rummelhart1988Learning}.
The remains usually take the form of building a new kind of recurrent unit on which the long-term dependence can be directly learned. For example,
Independent recurrent neural networks (IndRNN) does not allow any connections between different neurons, and also restricts each self-connection \cite{li2018independently}.
The IndRNN has the capability to handle long-term dependence that it can even work on the sequence longer than 5000 time steps.
Recently, there have also been efforts such as \cite{Quan2020External} to use additional mechanisms to assist RNN in learning long-term dependence.

Although these methods have achieved good performance in long sequence learning, they neglect the necessity to well learn the short-term dependence.
It also limits their usefulness on long-term dependence and prevents them from performing at their best.
This is very similar to the conflict problem presented in \cite{Wang2018Recurrent}, that recurrent units trained in the usual way cannot respond very well to an input or an output at a given time step while preserving memory.
Most of the gate-free networks cannot respond very well to short-term dependence while ensuring long-term dependence learning.
Consequently, the training process for these networks is usually slow.
Our model DuRNN follows the gate-free research line and combines advantages of all these previous methods to learn the sequential dependence.

\subsection{Stacked Models}

In the design of CNN, multiple convolutional layers are often stacked to construct a deep network for better performance.
For instance, VGG \cite{Simonyan2015Very} uses stacked convolution layers and alternated maximum pooling layers.
ResNet \cite{He2016Deep} uses a residual module to form a more complex network (network within network), which can be trained using the standard stochastic gradient descent method.

RNN can be also stacked into a multi-layer and deep networks.
The gated RNNs like HyperLSTM \cite{Ha2016HyperNetworks} can construct a multi-layer network so that the weights to generate gates can be affected by the dependence of the previous layer.
Hierarchical Multi-scale LSTM \cite{chung2017hierarchical} constructs a multi-scale network with several layers of LSTM of different gate units to further separate the different scales of dependence.
The gate-free RNNs as IndRNN can be stacked to a multi-layer network directly or be stacked to a residual recurrent network.
Both deep CNNs and RNNs can be used to process complex computer vision and scenes like \cite{LiAdaptive} and \cite{LiAccurate}.
Multi-layer RNNs have high research value and a wide range of applications.

\section{The Proposed Model}

Dual Recurrent Neural Networks (DuRNN) can be described with the following three equations,
\begin{eqnarray}
&\widetilde{\bm{h}}_{t}=\sigma_{f}({\bm{W}_{in}}\bm{x}_{t}+C_{\delta}({\bm{W}_{rec}})\widetilde{\bm{h}}_{t-1}+\bm{b}_{short}),\label{short_term}\\
&\bm{i}_{t}={\bm{S}_t\circ\widetilde{\bm{h}}_{t}},\label{selection}\\
&\bm{h}_{t}=\sigma_{s}(\bm{W}_{s}\bm{i}_{t}+{\bm{U}}\circ\bm{h}_{t-1}+\bm{b}_{long}),\label{long_term}
\end{eqnarray}
where $\bm{x}_{t}\in\mathbb{R}^M$ represents the input sequence, and $\bm{i}_{t}$ represents the intermediate output and input.
${\bm{W}_{in}}\in\mathbb{R}^{N\times{M}}$, $\bm{W}_{s}\in\mathbb{R}^{N\times{N}}$, ${\bm{W}_{rec}}\in\mathbb{R}^{N\times{N}}$ and $\bm{U}\in\mathbb{R}^{N}$ are the input matrices and the recurrent weight matrices for different recurrent connection parts, respectively.
$\sigma_{f}$ and $\sigma_{s}$ are the activation functions of the two parts, and here they are relu to be specific.
$\widetilde{\bm{h}}_{t},\bm{h}_{t}\in\mathbb{R}^{N}$ are the hidden states capturing the short-term and long-term memory, and $\bm{b}_{short},\bm{b}_{long}\in\mathbb{R}^N$ are the biases with respect to two recurrent connection parts, respectively.
$C_{\delta}$ is a clipping operation to control the singular values of a matrix which is similar to the model \cite{kanai2017preventing} and the symbol '$\circ$' is the Hadamard product.
The function $\bm{S}_t$ is a selection mechanism of channel which selects the short-term information to be transferred into the long-term memory. It is set as follows:
\begin{eqnarray}
\bm{S}_t = relu(mm({\bm{W}_{ss}}\widetilde{\bm{h}}_{t}+{\bm{W}_{ls}}\bm{h}_{t-1}+\bm{b}_{s})-\bm{b_{thre}}),\label{generate_St}
\end{eqnarray}
where $mm$ is a channel response normalization, which uses the min-max normalization function to normalize the vector by a linear operation:
\begin{eqnarray}
mm(vector) = \frac{vector - min(vector)}{max(vector) - min(vector)}.\label{min_max}
\end{eqnarray}
${\bm{W}_{ss}}\in\mathbb{R}^{N\times{N}}$, ${\bm{W}_{ls}}\in\mathbb{R}^{N\times{N}}$ and $\bm{b}_{s}\in\mathbb{R}^{N}$ are the weight matrix and bias to generate the selection vector according to long-term and short-term memory.
$\bm{b}_{thre}\in\mathbb{R}$ is a scalar variable used as a threshold to control the transmitting of information, and it is in the interval of [0,1].
$M$ and $N$ are the dimensions of the input feature and the number of neurons in the hidden states, respectively.

Fig. 1 shows the framework of DuRNN, which can be divided into two parts and a selection mechanism. The left part is in charge of separating short-term dependence which corresponds to Eq.(\ref{short_term}). The sequence input initially disperses its natural features to the preset neurons through the input matrix ${\bm{W}_{in}}$. Then, the neurons interact with each other by the recurrent weight matrix ${\bm{W}_{rec}}$ and give the corresponding outputs with the activation function relu over time. We further clip the singular values of the recurrent weight matrix ${\bm{W}_{rec}}$  below a threshold value $\delta$ ($|\delta|<1$) to only learn the short-term dependence. As reported in \cite{pascanu2013difficulty} and \cite{Bengio1993The}, it is hard to capture long-term information. Short-term and periodic dependence can be learned and the non-sequential information like noise will be discarded soon in this part.
It is worth mentioning that the $\delta$ needs to be set appropriately so that short-term dependence can be well learned without affecting the capture of long-term dependence.
If $\delta$ is set as 1, the long-term and short-term dependence can not be separated well, which will affect the performance of training.
This point will be confirmed in the experiment section.

\begin{figure*}[t]
	\begin{center}
		\includegraphics[scale=0.4]{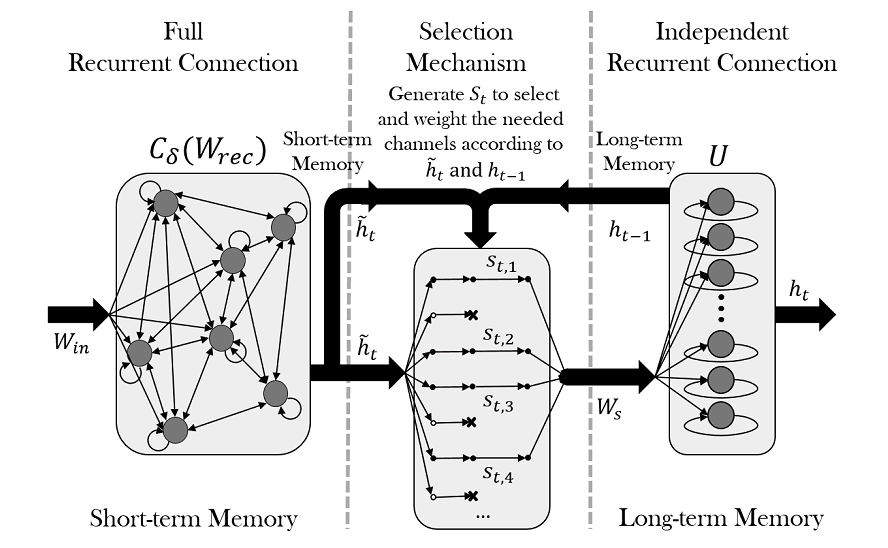}
	\end{center}
	\caption{The framework of DuRNN. It consists of two parts and a transition connection. The left and right parts in the figure correspond to the short-term and long-term memories respectively. The middle part is the transition connection which optimizes the flow of information. The selection weights are generated from the short-term memory of current time step and the long-term memory of previous time step. }
\end{figure*}

Then, the state with the short-term dependence information comes into a selection mechanism introduced in Eq.(\ref{selection}) and shown in the middle of Fig.1.
The short-term state and long-term state are integrated and normalized, then clipped by the trainable threshold $b_{thre}$, and finally go through the relu function and generate the selection weight $\bm{S}_t$ as in Eqs.(\ref{generate_St}-\ref{min_max}).
And in Eq.(\ref{selection}), the information in short-term state can be selected and paid varying degrees of attention by $\bm{S}_t$ according to the current memory state.

The right part of Fig.1 corresponds to Eq.(\ref{long_term}), which helps to learn the long-term dependence.
The selected short-term memory is reweighted by the input weight matrix $\bm{W}_s$ to reorganize its spatial information because each neuron has a different global importance, and then enter the independent connection.
The idea of independent connection is proposed in \cite{li2018independently}.
Since there do not exist inter-connections, the gradient value on each neuron changes very smoothly with the help of controlling $\bm{U}$. By this property, the long-term dependence can be well learned on these neurons when the recurrent weight vector $\bm{U}$ is close to the unit vector.

DuRNN can be naturally stacked to a multi-layer network.
The outputs of DuRNN are formed as a sequence entering the next layer as the input.
The structure of DuRNN and the selection mechanism can help the multi-layer recurrent neural network to capture the hierarchical sequence dependence adaptively, and learn the sequential dependence information of different scales separately.

The divide-and-conquer strategy empowers DuRNN the capability to capture long-term dependence quickly and accurately.
Unlike LSTM, DuRNN does not use forget gates to manipulate long-term memory, so it can avoid the problem of gradient vanishing caused by the forgetting mechanism of general gated RNNs.
Different from IndRNN \cite{li2018independently}, which trys to learn short-term dependence with a loose constrained long-term model, DuRNN has a dedicated network structure for short-term memory.
So the independent neurons do not need to reduce their recurrent weights to learn short-term dependence, which makes the optimization process more directed, and the training faster.
That also allows more independent neurons to be activated to capture long-term dependence, therefore the performance of independent connections will be improved.

The selection mechanism is an essential part of DuRNN to implement divide-and-conquer strategy and consolidate its improvement in different situations.
The most important function of this mechanism is to control the accumulation and flow of information between them according to the current long-term and short-term memory.
This selection mechanism replaces the traditional input matrix or gates to transmit information, and help DuRNN to process the dependence in sequences selectively.
The motivation of selection mechanism is like the local response normalization (LRN) \cite{krizhevsky2012imagenet}, which is designed to inhibite some neurons and relatively highlight the others, but it is different from LRN in method.
One is the selection of channel is adaptive, determined by the current memory state, and does not specify the select location.
The other is that elements in $S_t$ generated by $mm$ and $b_{thre}$ are either 0 or between 0 and 1, which makes the selection more radical.

\section{Training and Analysis}
\subsection{Gradient Back Propagation}

In this section, we will follow the route of BPTT \cite{Rummelhart1988Learning}. As stated in \cite{pascanu2013difficulty}, for gradient back propagation, the following terms should  be calculated,
\begin{eqnarray}
&\displaystyle\frac{\partial{Loss}}{\partial{\theta}}=\sum_{1\leq i\leq L}\frac{\partial{Loss_i}}{\theta},\label{bptt_all}\\
&\displaystyle\frac{\partial{Loss_i}}{\partial{\theta}}= \sum_{1\leq j\leq i}\frac{\partial{Loss_i}}{\partial{h_{i}}}\frac{\partial{h_i}}
{\partial{h_{j}}}\frac{\partial{h}_{j}}{\partial{\theta}},\label{btpp_t}\\
&\displaystyle\frac{\partial{h_i}}{\partial{h_{j}}} = \prod_{j+1\leq k \leq i}\frac{\partial{h_k}}{\partial{h_{k-1}}},\label{bptt_prod}
\end{eqnarray}
where $Loss_i$ is the loss at the $i$th time step, the total $Loss$ equals to $\sum_i{Loss_i}$, $\theta$ represents the learnable variables and $L$ is the length of input sequence.

As shown in Eq.(\ref{bptt_prod}), we first need to calculate the gradients $\partial{h_t}/\partial{h_{t-1}}$. From Eq.(\ref{long_term}), the following equations will be derived,
\begin{eqnarray*}
\frac{\partial{h}_{t}}{\partial{h}_{t-1}}=diag(U\circ \sigma_{s,t}^{'})+\frac{\partial{h}_{t}}{\partial{i}_{t}}\frac{\partial{i}_{t}}
{\partial{S}_{t}}\frac{\partial{S}_{t}}{\partial{h}_{t-1}},
\end{eqnarray*}
where $\sigma_{s,t}^{'}(\cdot)$ is the derivative of the activation function $\sigma_s(\cdot)$ at time $t$.
However, the above derivative has the term of the selection part. This term may interfere the control of $U$ and affect the long-term dependence learning, so we force $\partial{S}_{t}/\partial{h}_{t-1}$ to be zero in the training process.
To simplify the calculation, the term $\partial{S}_{t}/\partial{\widetilde{h}}_{t}$ is also truncated in the following derivative,
\begin{eqnarray*}
\displaystyle\frac{\partial{h}_t}{\partial{\widetilde{h}_{t}}}=W_{s}^Tdiag(S_{t}\circ \sigma_{s,t}^{'})+\frac{\partial{h}_{t}}{\partial{i}_{t}}\frac{\partial{i}_{t}}
{\partial{S}_{t}}\frac{\partial{S}_{t}}{\partial{\widetilde{h}_{t}}}.
\end{eqnarray*}
Thus, the gradient back propagation between adjacent states are simplified as follows,
\begin{eqnarray}
\displaystyle\frac{\partial{h}_{t}}{\partial{h}_{t-1}}&=&diag(U\circ \sigma_{s,t}^{'}), \label{g_long_to_long} \\
\displaystyle\frac{\partial{h}_t}{\partial{\widetilde{h}_{t}}}&=&diag(S_{t})W_{s}^Tdiag(\sigma_{s,t}^{'}),\label{g_long_to_short}\\
\displaystyle\frac{\partial{\widetilde{h}_{t}}}{\partial{\widetilde{h}_{t-1}}}
&=&C_\delta(W_{rec})^Tdiag(\sigma_{f,t}^{'}),\label{g_short_to_short}
\end{eqnarray}
where $\sigma_{f,t}^{'}(\cdot)$ is the derivative of the activation function $\sigma_f(\cdot)$ at time $t$.
In this case, $S_t$ becomes a selection mechanism based on the current state.
For simplicity, $\partial{h}_t/\partial{\widetilde{h}}_{t}$ is denoted as $dS_t$.

For computing the derivatives of loss with respect to the state variables $h$ and $\widetilde{h}$, based on Eqs.(\ref{bptt_all}-\ref{bptt_prod}), we have the following basic fact,
\begin{eqnarray}
\frac{\partial{Loss}}{\partial{h_t}}&=&\sum_{0<i<L+1}\!\!\frac{\partial{Loss_i}}{\partial{h_t}}
=\sum_{t-1<i<L+1}\!\!\frac{\partial{Loss_i}}{\partial{h_t}}\nonumber\\
&=&\frac{\partial{Loss_t}}{\partial{h_t}}+\sum_{t<i<L+1}\!\!\frac{\partial{Loss_i}}
{\partial{h_{t+1}}}\frac{\partial{h_{t+1}}}{\partial{h_t}}\nonumber\\
&=&\frac{\partial{Loss_t}}{\partial{h_t}}+\frac{\partial{Loss}}{\partial{h_{t+1}}}
\frac{\partial{h}_{t+1}}{\partial{h}_{t}}.\label{bptt_iterate}
\end{eqnarray}
Thus, the following gradients can be derived,
\begin{eqnarray}
&\displaystyle\frac{\partial{Loss}}{\partial{h_t}}\!=\!\displaystyle\frac{\partial{Loss_t}}{\partial{h_t}}\!+\!diag(U\circ \sigma_{s,t+1}^{'})\frac{\partial{Loss}}{\partial{h_{t+1}}},\label{g_long}\\
&\!\!\!\!\!\displaystyle\frac{\partial{Loss}}{\partial{\widetilde{h}_{t}}}\!=\!
dS_{t}\displaystyle\frac{\partial{Loss}}{\partial{h_t}}\!+\!C_\delta(W_{rec})^Tdiag(\sigma_{f,t+1}^{'})\frac{\partial{Loss}}{\partial{\widetilde{h}_{t+1}}}.\label{g_short}
\end{eqnarray}

For the gradient with respect to the weight $U$ of long-term memory, it can be calculated as
\begin{eqnarray*}
&\displaystyle\frac{\partial{Loss}}{\partial{U}}=\sum_tdiag(\sigma_{s,t}^{'})
\displaystyle\frac{\partial{Loss}}{\partial{{h}_{t}}}\circ {h}_{t-1}.
\end{eqnarray*}

For the gradients with respect to the weights of short-term memory, they are
\begin{eqnarray*}
&\displaystyle\frac{\partial{Loss}}{\partial{W_{rec}}}=
\sum_tdiag(\sigma_{f,t}^{'})\displaystyle\frac{\partial{Loss}}{\partial{\widetilde{h}_{t}}}\widetilde{h}_{t-1}^T,\\
&\displaystyle\frac{\partial{Loss}}{\partial{W_{in}}}=
\sum_tdiag(\sigma_{f,t}^{'})\displaystyle\frac{\partial{Loss}}{\partial{\widetilde{h}_{t}}}\widetilde{x}_{t}^T.
\end{eqnarray*}

Similarly, the gradients with respect to $W_s$, $W_{ss}$, $W_{ls}$ and $b_{thre}$ in the selection mechanism are
\begin{eqnarray*}
&\!\displaystyle\frac{\partial{Loss}}{\partial{W_{s}}}\!=\!\sum_t(\sigma_{s,t}^{'}\circ \frac{\partial{Loss}}{\partial{h_t}})(S_{t}\circ \widetilde{h}_{t})^T,\\
&\!\displaystyle\frac{\partial{Loss}}{\partial{W_{ss}}}\!=\!\sum_t (W_{s}^T(\sigma_{s,t}^{'}\circ {\frac{\partial{Loss}}{\partial{h_t}})\circ \widetilde{h}_{t}\circ relu^{'}\circ mm^{'}})\widetilde{h}_{t}^T,\\
&\!\!\displaystyle\frac{\partial{Loss}}{\partial{W_{ls}}}\!=\!\sum_t (W_{s}^T(\sigma_{s,t}^{'}\circ {\frac{\partial{Loss}}{\partial{h_t}})\circ \widetilde{h}_{t}\circ relu^{'}\circ mm^{'}})h_{t-1}^T,\\
&\!\!\!\!\displaystyle\frac{\partial{Loss}}{\partial{b_{thre}}}\!=\!-\sum_t\sum_{vector} W_{s}^T(\sigma_{s,t}^{'}\circ {\frac{\partial{Loss}}{\partial{h_t}})\circ \widetilde{h}_{t}\circ relu^{'}},
\end{eqnarray*}
where $relu^{'}$ and $mm^{'}$ are the derivatives of corresponding functions in Eq.(\ref{generate_St}).
The symbol $\sum\nolimits_{vector}$ means the sum of all elements in the vector.
The gradient for bias can be calculated iteratively in the similar way. For detailed computation, please refer to Appendix.

\subsection{Gradient Control}

As shown in the above section, all parameters can be learned end-to-end. Now we will show the gradient vanishing will not happen on the independent connections and the gradient exploding will not happen at all.

First of all, by constraining the weight $U$,  the gradient vanishing problem will not happen most of the time.
To illustrate it, the $L$-step back propagation ${\partial{h}_{t}}/{\partial{h}_{t-L}}$ is calculated according to Eq.(\ref{g_long}),
\begin{eqnarray*}
\frac{\partial{h}_{t}}{\partial{h}_{t-L}}&=&\prod_{k=t-L+1}^{t}diag(U\circ \sigma_{s,k}^{'}).
\end{eqnarray*}
Since all elements in the vector $U$ are kept within the interval $[\sqrt[L]{\epsilon},\sqrt[L]{\gamma}]$ $(0<\epsilon<1<\gamma)$, ${\partial{h}_{t}}/{\partial{h}_{t-L}}$ can be constrained in the following interval,
\begin{eqnarray*}
\epsilon\left\|\prod_{k=t-L+1}^{t}\!\!\!\!diag(\sigma_{s,k}^{'})\right\|<\left\|\frac{\partial{h}_{t}}{\partial{h}_{t-L}}\right\|<\gamma\!\!\!\!\prod_{k=t-L+1}^{t}\!\left\|diag(\sigma_{s,k}^{'})\right\|.
\end{eqnarray*}
Because the active function is relu, the elements in $\prod_{k=t-L+1}^{t}\sigma_{s,k}^{'}$ are either 0 or 1. If the networks are initialized properly, the elements in $\prod_{k=t-L+1}^{t}\sigma_{s,k}^{'}$ will be one. Thus there exists a lower bound $\epsilon$. It means that the gradients are kept from vanishing in $L$ time steps. This property helps the independent connections capture the dependence of long-term sequence, which is shown in \cite{li2018independently}.
The characteristics of RNN with full connections enable it to learn the short-term dependence, while the gradient back propagation on it cannot be guaranteed.



The gradient explosion can also be prevented. From Eqs.(\ref{g_long_to_long}-\ref{g_short_to_short},\ref{g_long}-\ref{g_short}), the gradient back propagation from $h_t$ to $\widetilde{h}_{t-L}$ can be calculated as
\begin{equation*}
\begin{split}
&\frac{\partial{h}_t}{\partial{\widetilde{h}_{t-L}}}=\\
&\sum_{i=0}^{L}(\prod_{n=t-L}^{t-i-1}(C_\delta(W_{rec})^Tdiag(\sigma_{f,n}^{'}))dS_{t-i}
\prod_{m=t-i}^{t-1}diag(U\circ \sigma_{s,m}^{'})),
\end{split}
\end{equation*}
where $L$ the length of the sequence. Because the derivative of relu $\sigma^{'}\le 1$, the following inequality will holds,
\begin{equation*}
\left\|\frac{\partial{h}_t}{\partial{\widetilde{h}_{t-L}}}\right\| \le\sum_{i=0}^{L}\left\| dS\right\|{\left\|  diag(U)\right\|}^i{\left\| C_\delta(W_{rec})^T\right\|}^{L-i},
\end{equation*}
where the norm is the $L_2$ norm and $\|dS\|=s$ is the upper bound of $\|dS_t\|$. Then it can be shown that the gradient is under control,
\begin{equation*}
\left\|\frac{\partial{h}_t}{\partial{\widetilde{h}_{t-L}}}\right\|
<s\gamma\sum_{i=0}^{L}\delta^{L-i}<\frac{s\gamma}{1-\delta},
\end{equation*}
where $\gamma$ and $\delta$ are all independent of the length $L$, so the gradient of back propagation has an upper bound globally.

\section{Experiments}

\subsection{Visualization Section}

In this section we use the addition problem to compare DuRNN with several classical RNN models and carry out the ablation analysis of DuRNN. We also show the neuron activations of trained DuRNN and its characteristics of learning sequential dependence to rationalize our design.

\textbf{The adding problem} \cite{hochreiter1997long} is to evaluate a RNN model by a sequential task.
A two-dimensional sequence is given as input.
Each element in the sequence has two features.
The first is a random number in the range $(0, 1)$ .
There are only two elements in the sequence in which the second feature is 1, and all the others are 0.
The problem is to compute the sum of the first features of two elements in which the second feature is 1.
Each input is generated randomly and the correct result is calculated as label in advance.

In the adding problem, the standard RNN with tanh active function, LSTM \cite{hochreiter1997long}, IRNN \cite{le2015simple} and IndRNN \cite{li2018independently} are chosen for comparisons.
RNN, LSTM, IRNN and DuRNN are all one-layer networks while the IndRNN is a two-layer network.

The sequence length $L$ will be set as 100, 500, 1000 and 5000 to evaluate the performance of these models.
The initial learning rate is uniformly set as $2\times10^{-4}$, and decreases to 0.1 times of the original after each 20000 iterations in the training.
It can be set larger on learning rate, but since IRNN and RNN have no singular value constraint and are sensitive to the changes of learning rate, too large learning rate may cause instability in the training process, so we kept it at a relatively small value.
128 neurons are used in each recurrent sublayer for all models and the batch size is 50.
The parameters $\epsilon$ and $\gamma$ are set as 0.5 and 2 and the parameter $\delta$ is usually set as $\sqrt[L]{0.5}$.
The task of the adding problem uses the mean squared error (MSE) as its loss function and Adam optimizer \cite{Kingma2015Adam} to train the model.
In Fig. 2, the error descent curves are drawn in the logarithmic coordinate, so that we can have a clearer comparison of the training process between all models. For the sequences with lengths 1000 and 5000,  the models LSTM and IRNN do no work. So their curves in these cases are not drawn.
If the RNN model does not learn anything, the MSE will be 0.167, which is the variance of the sum of two independent uniform distributions. In this case, the error curve will oscillate around this value.

\begin{figure*}
	\begin{center}
		\begin{subfigure}[b]{0.32\textwidth}
			\includegraphics[width=\textwidth]{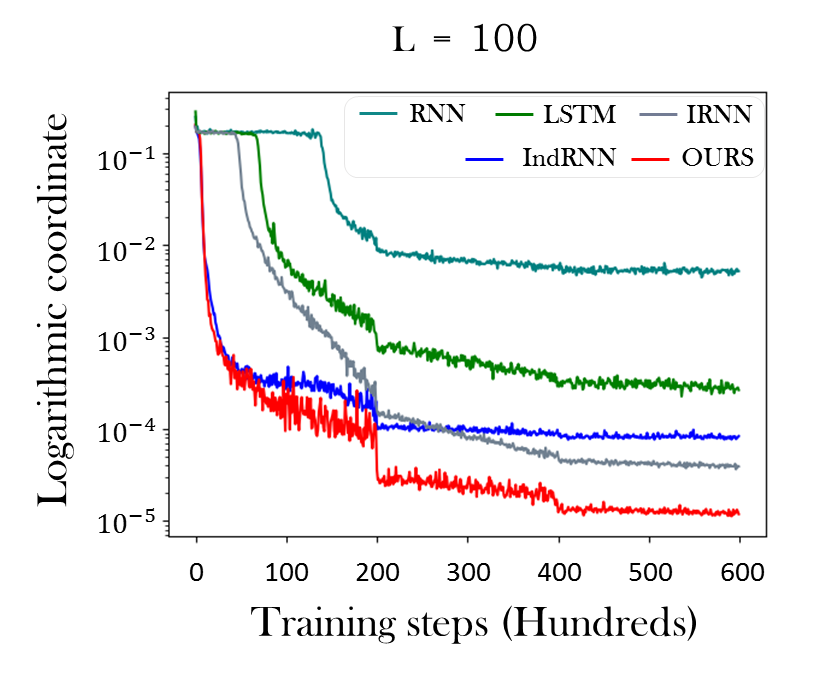}
			\caption{}
		\end{subfigure}
        \qquad
		\begin{subfigure}[b]{0.32\textwidth}
			\includegraphics[width=\textwidth]{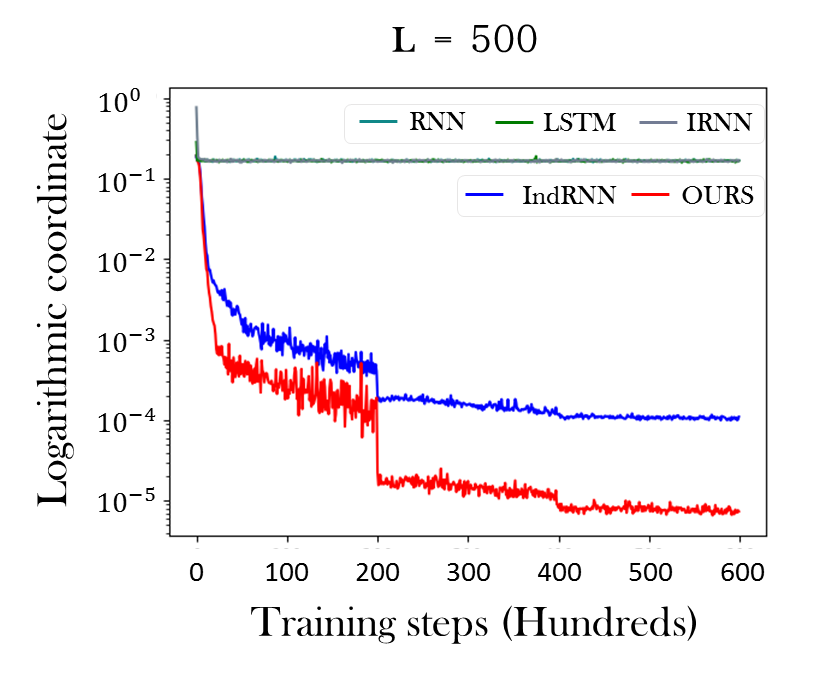}
			\caption{}
		\end{subfigure}
	\end{center}
	\begin{center}
		\begin{subfigure}[b]{0.32\textwidth}
			\includegraphics[width=\textwidth]{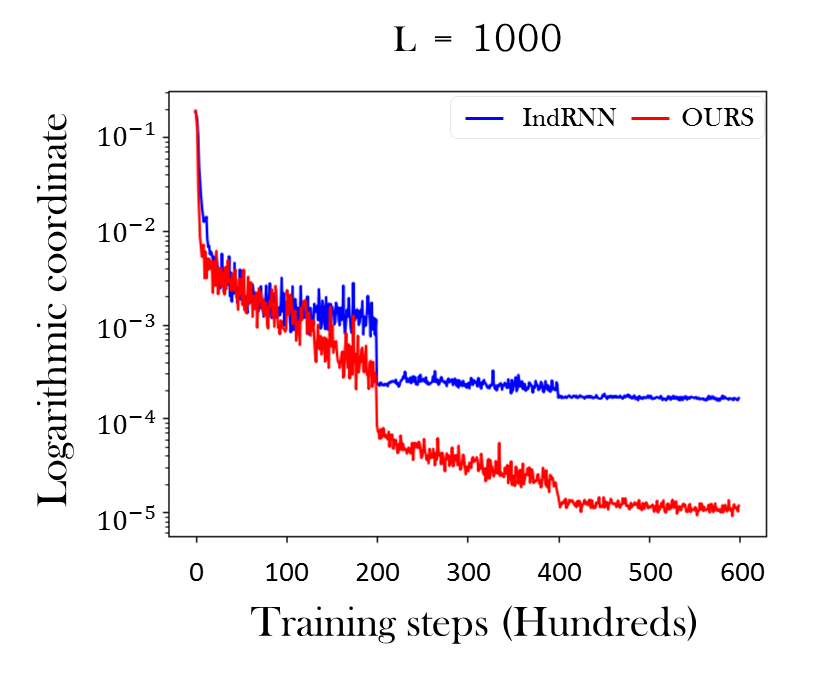}
			\caption{}
		\end{subfigure}
        \qquad
		\begin{subfigure}[b]{0.32\textwidth}
			\includegraphics[width=\textwidth]{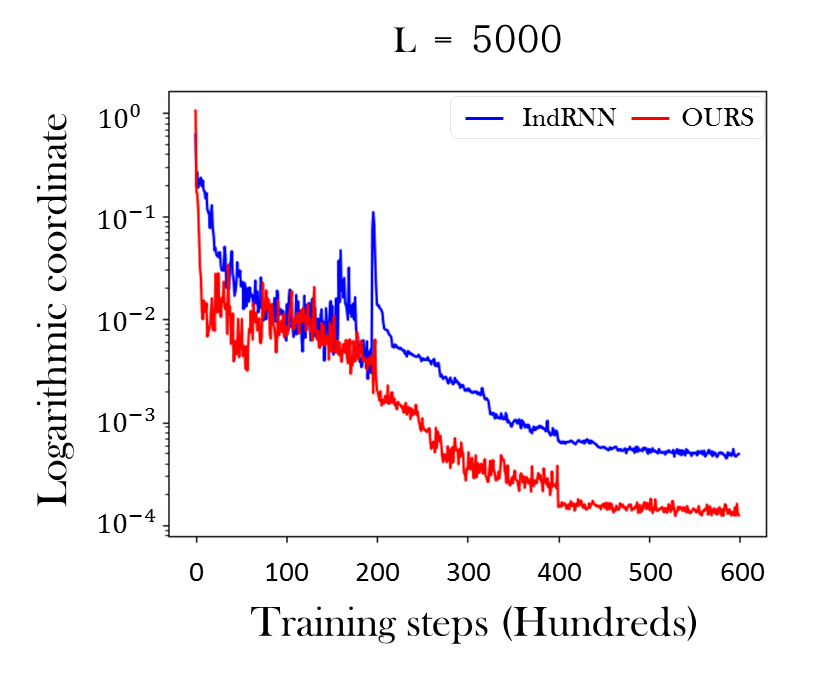}
			\caption{}
		\end{subfigure}
	\end{center}
	\caption{The error descent curves of different RNNs in the adding problem. The sequences with 100, 500, 1000 and 5000 steps are generated in real time and used for training and testing. The abscissa represents the number of training steps, and the ordinate is the logarithmic mean square error. }
\end{figure*}

As shown in Fig. 2, DuRNN shows the capability to handle long sequence and perform well over thousands of steps.
Furthermore, we can also see that DuRNN has the steepest and smoothest descent curve at the same learning rate and obtains higher accuracy.
This indicates that our model is more rapid, accurate and noise resistant for sequence dependence learning.
In the section of ablation analysis later, we will specifically analyze the effects brought by each part of DuRNN.

\subsubsection{Ablation Analysis}

\begin{figure*}
	\begin{center}
		\begin{subfigure}[b]{0.32\textwidth}
			\includegraphics[width=\textwidth]{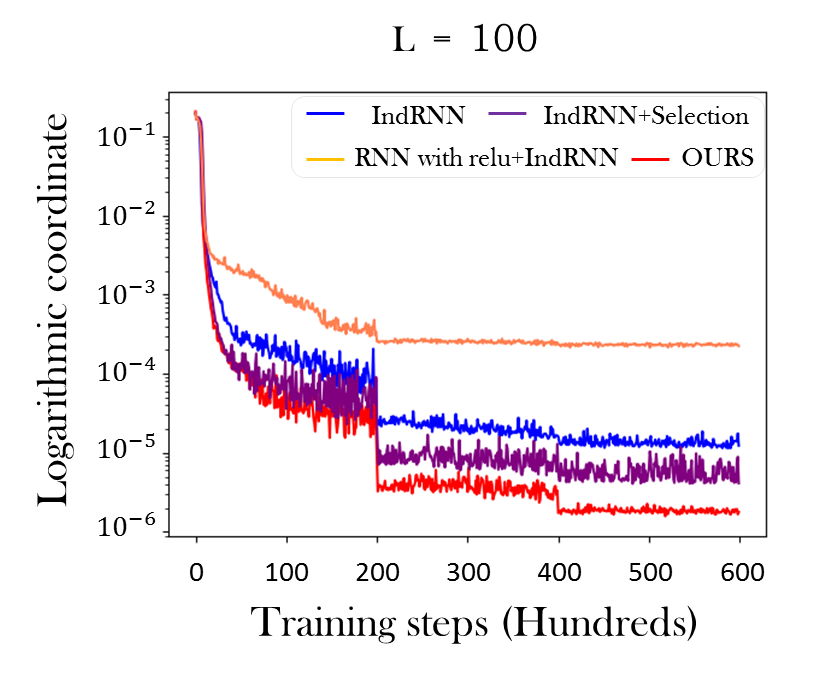}
			\caption{}
		\end{subfigure}
        \qquad
		\begin{subfigure}[b]{0.32\textwidth}
			\includegraphics[width=\textwidth]{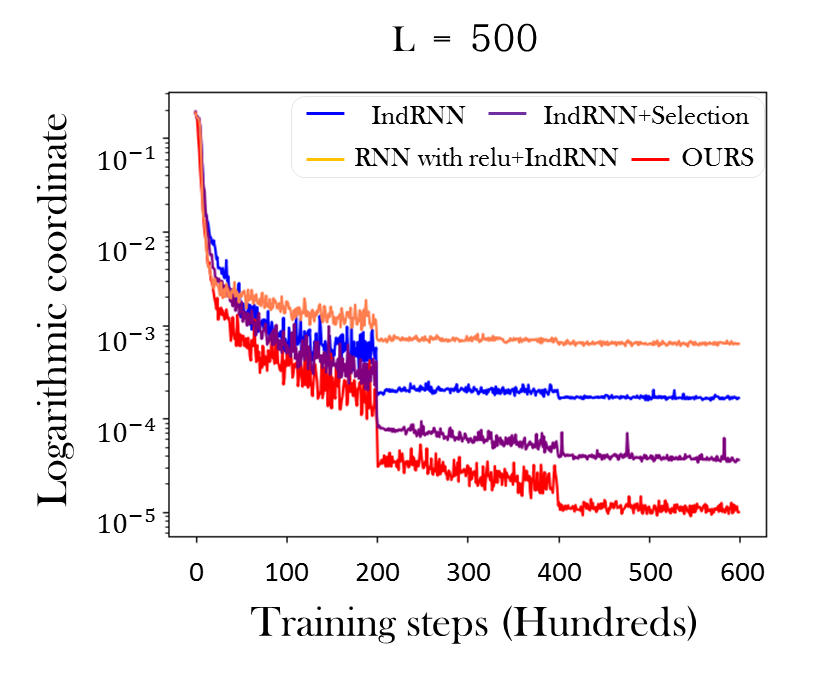}
			\caption{}
		\end{subfigure}
	\end{center}
	\begin{center}
		\begin{subfigure}[b]{0.32\textwidth}
			\includegraphics[width=\textwidth]{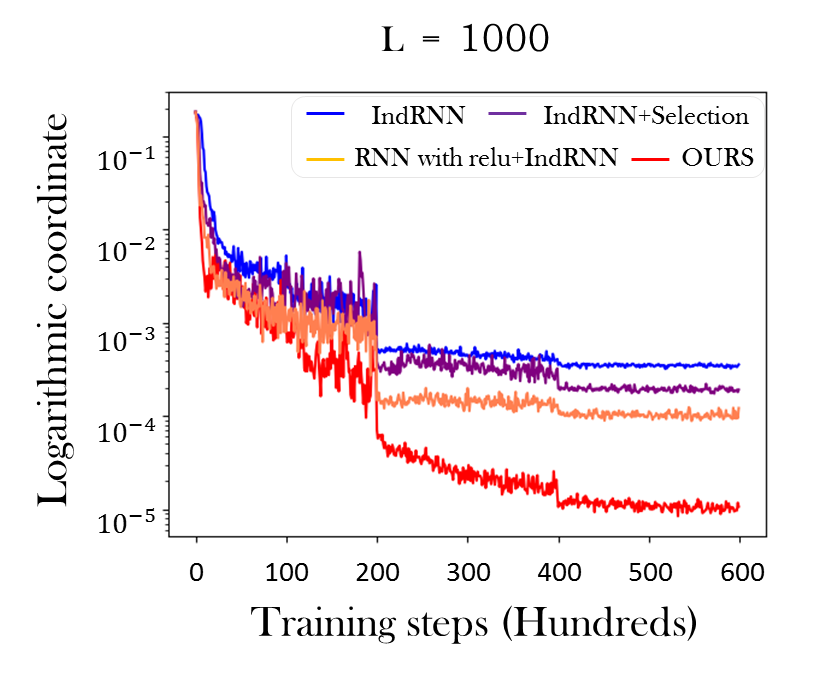}
			\caption{}
		\end{subfigure}
        \qquad
		\begin{subfigure}[b]{0.32\textwidth}
			\includegraphics[width=\textwidth]{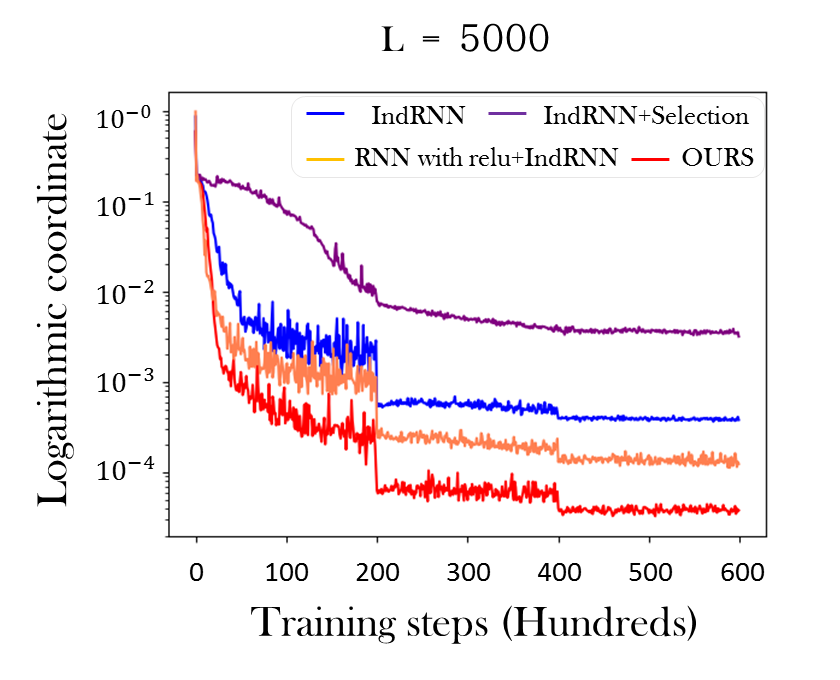}
			\caption{}
		\end{subfigure}
	\end{center}
	\caption{The ablation experiments of DuRNN. The settings is the same as that in Fig. 1. DuRNN without divide-and-conquer strategy and DuRNN without selection mechanism are compared with IndRNN and DuRNN itself.}
\end{figure*}

We set ablation experiments in the task of adding problem, and the error descent curves of different models are also shown in Fig. 3.
RNN with relu + IndRNN contains full and independent recurrent connections and IndRNN + Selection is a two-layer IndRNN with the selection part before the second layer.
The former can be viewed as a DuRNN without the selection part; the latter is a DuRNN in which recurrent connection of the first sublayer is replaced by independent recurrent connection.

In Fig. 3 we can clearly see the effects of the two kinds of connections and the selection part.
In the case of processing short sequences in Fig. 3(a) and Fig. 3(b), DuRNN without selection has a relatively bad performance.
Because RNN with relu can already capture the dependence of this length, the divide-and-conquer strategy fail and both the short and long-term information enters the independent neurons.
One-layer independent recurrent connection can hardly process it, so the performance is affected.
After adding the selection mechanism, the expression precision of the dependence can be improved.
And the selection mechanism on the basis of independent connection is a certain but limited improvement.

In the case of processing long or very long sequences over 1000 steps, as Fig. 3(c) and Fig. 3(d) shown, RNN with relu loses the capability to deal with dependence in long sequences, so it will focus on the preprocessing of the short-term information.
And we can observe that the divide-and-conquer strategy significantly improves the efficiency of training and the final results.
In contrast, the selection mechanism applied to IndRNN affects its performance.
This is because the independent connections can capture long-term dependence in both of the two layers, so the selection mechanism hinders the correct flow of information.

The divide-and-conquer strategy enables DuRNN to capture long-term dependence well, and the selection mechanism enables the divide-and-conquer strategy to be implemented correctly on short sequences.
Therefore, DuRNN has the best adaptability in different cases and can effectively and accurately capture the dependence in sequences of different lengths.

\subsubsection{Dependence Learning}

We set experiments to observe the neuron activations of different layers in the trained multi-layer RNN models.
Because the relu activation function is used, a neuron is regarded to be active when the output on it stay positive.
We plot the time-varying activations for all neurons in a layer, and these neuron activations can reveal whether the corresponding layer learns the long-term dependence.
Based on \cite{li2018independently}, if there are continuously active neurons in some layer, i.e. the outputs of these neurons stay non-zero, and their curves are relatively smooth, we think that this layer has learned some kind of long-term dependence. Meanwhile, if neuron activations in some layer changes suddenly over time, or the outputs stay zero, we think the layer captures only short-term dependence, or nothing.

\begin{figure*}
	\begin{center}
		\begin{subfigure}[b]{0.32\textwidth}
			\includegraphics[width=\textwidth]{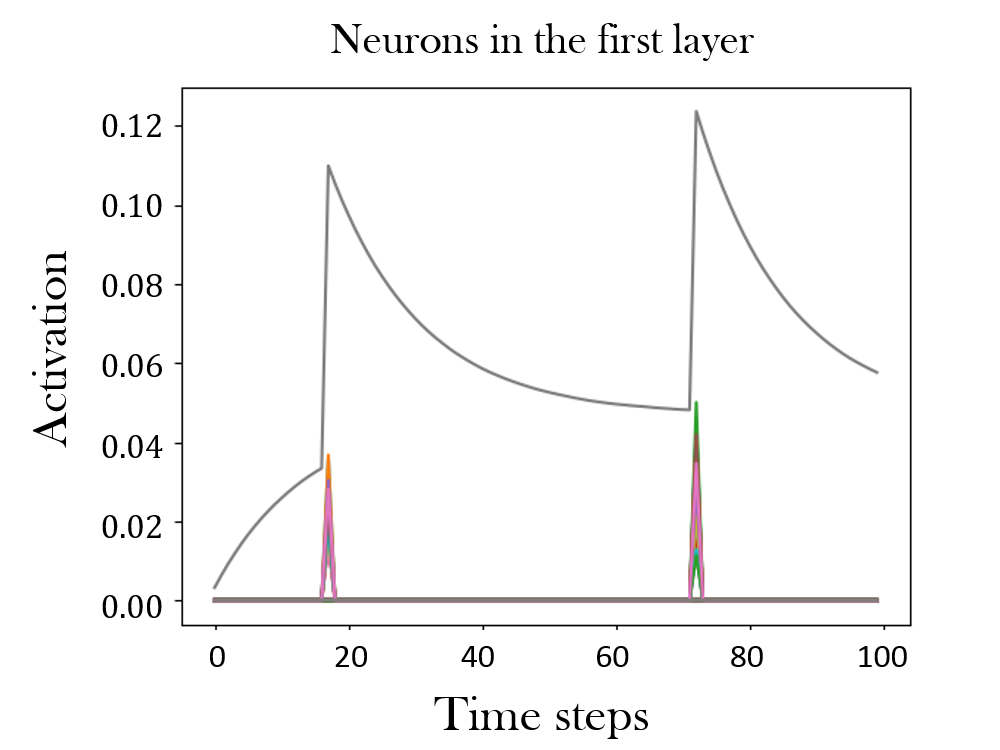}
			\caption{}
		\end{subfigure}
		\begin{subfigure}[b]{0.32\textwidth}
			\includegraphics[width=\textwidth]{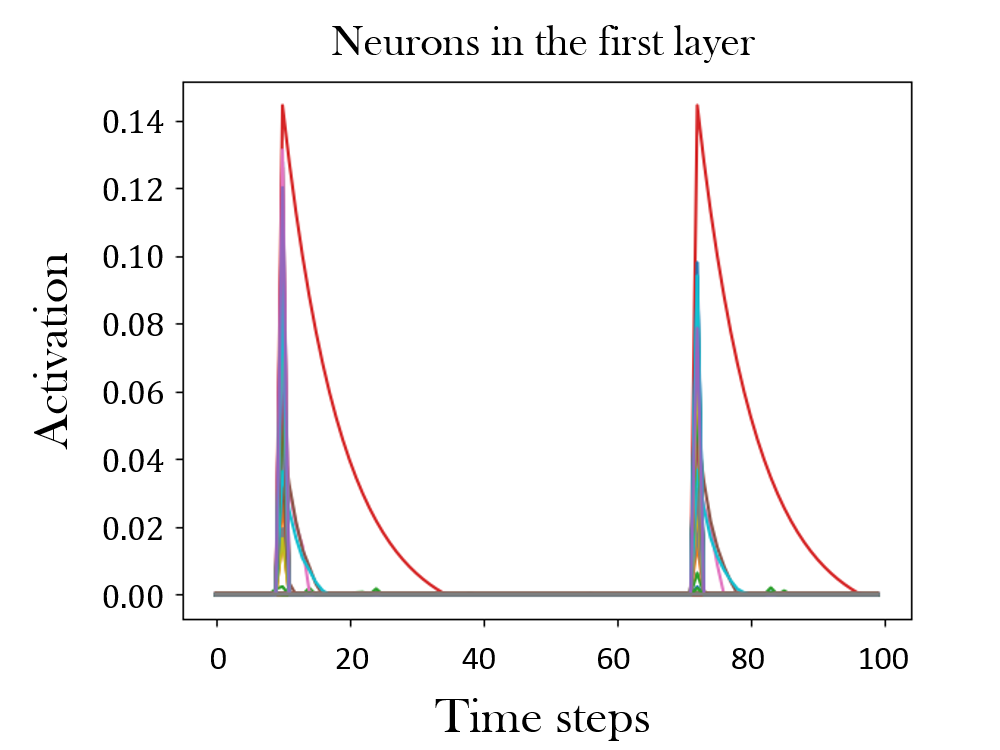}
			\caption{}
		\end{subfigure}
		\begin{subfigure}[b]{0.32\textwidth}
			\includegraphics[width=\textwidth]{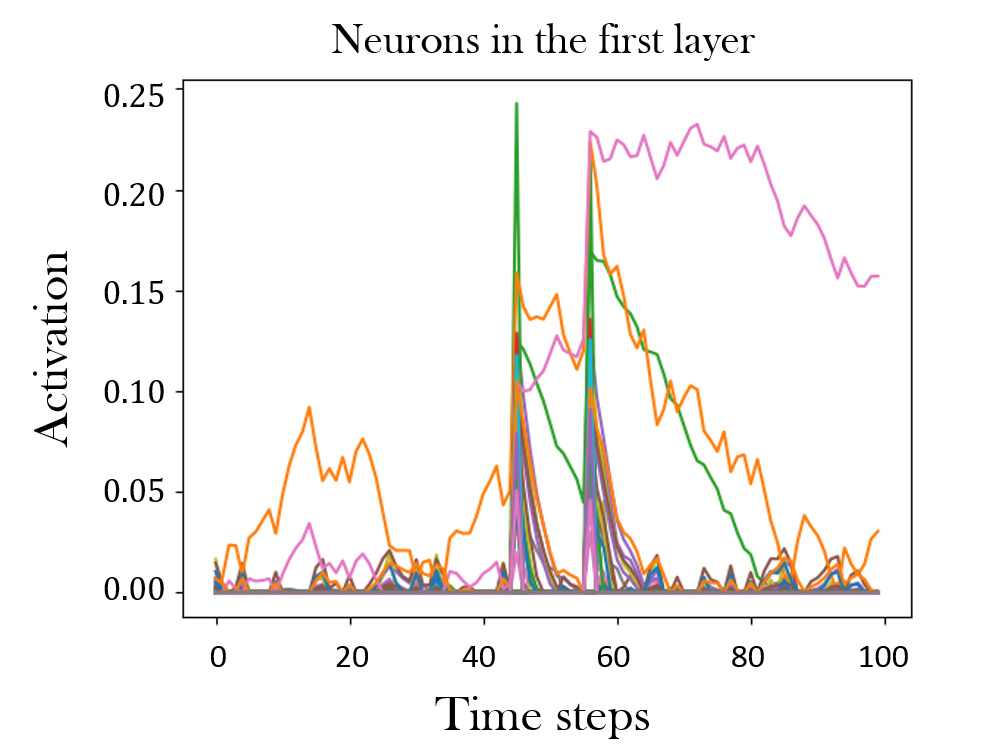}
			\caption{}
		\end{subfigure}
	\end{center}
	\begin{center}
		\begin{subfigure}[b]{0.32\textwidth}
			\includegraphics[width=\textwidth]{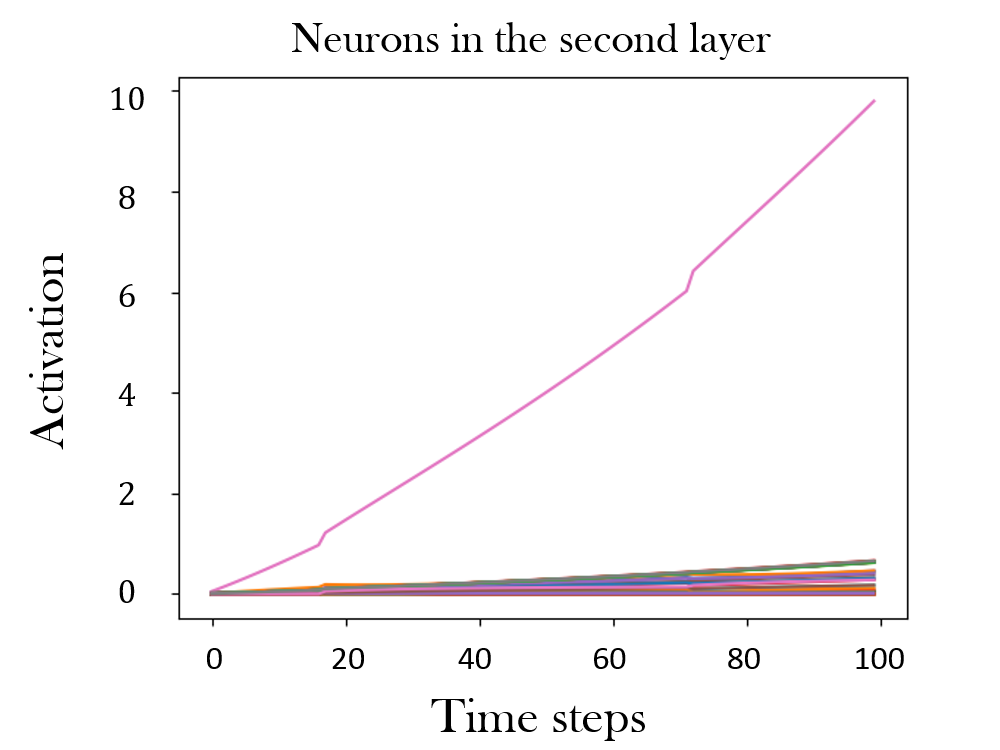}
			\caption{}
		\end{subfigure}
		\begin{subfigure}[b]{0.32\textwidth}
			\includegraphics[width=\textwidth]{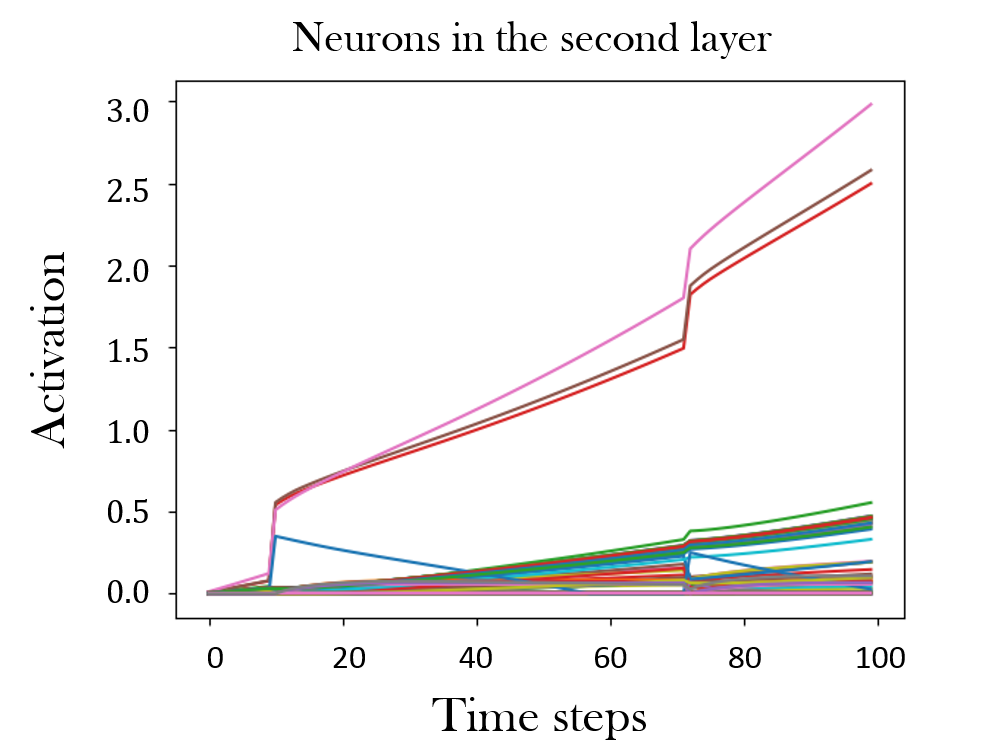}
			\caption{}
		\end{subfigure}
        \begin{subfigure}[b]{0.32\textwidth}
			\includegraphics[width=\textwidth]{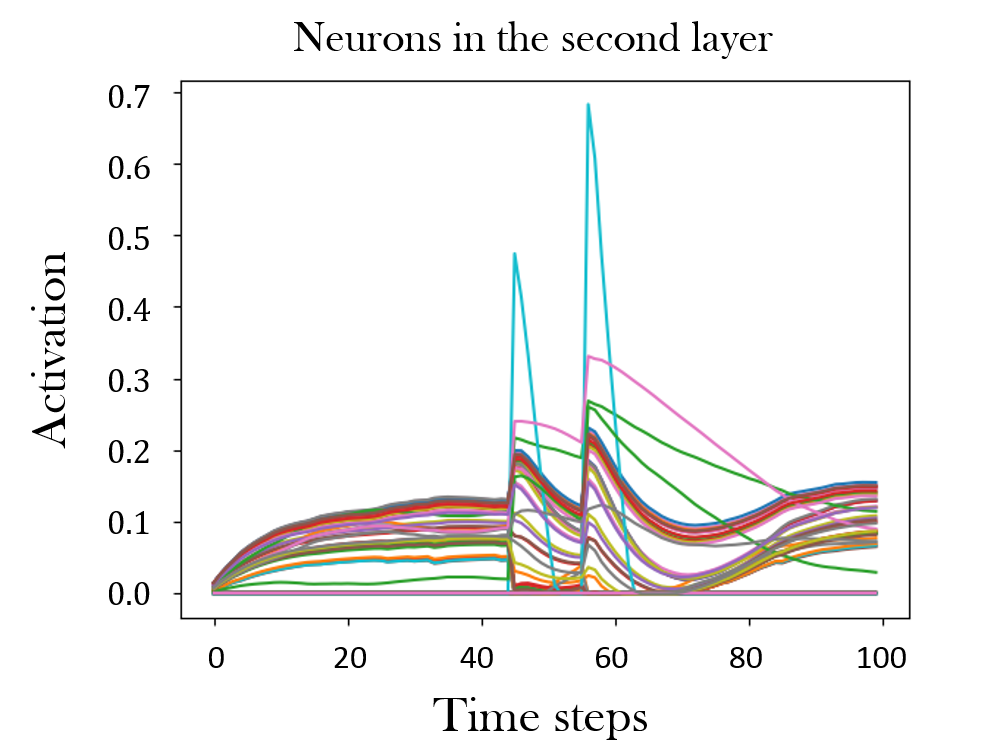}
			\caption{}
		\end{subfigure}
	\end{center}
	\caption{Neuron activations of 2-layer IndRNNs under different constraints. The figure shows neuron outputs in the first and second layers of different models, and each output curve represents the activations of a single neuron. The abscissa represents the activation value, and the ordinate is the time steps of the sequence. The activations of neuron can reveal whether the corresponding layer learns long-term dependence.}
\end{figure*}

\begin{figure*}
	\begin{center}
		\begin{subfigure}[b]{0.32\textwidth}
			\includegraphics[width=\textwidth]{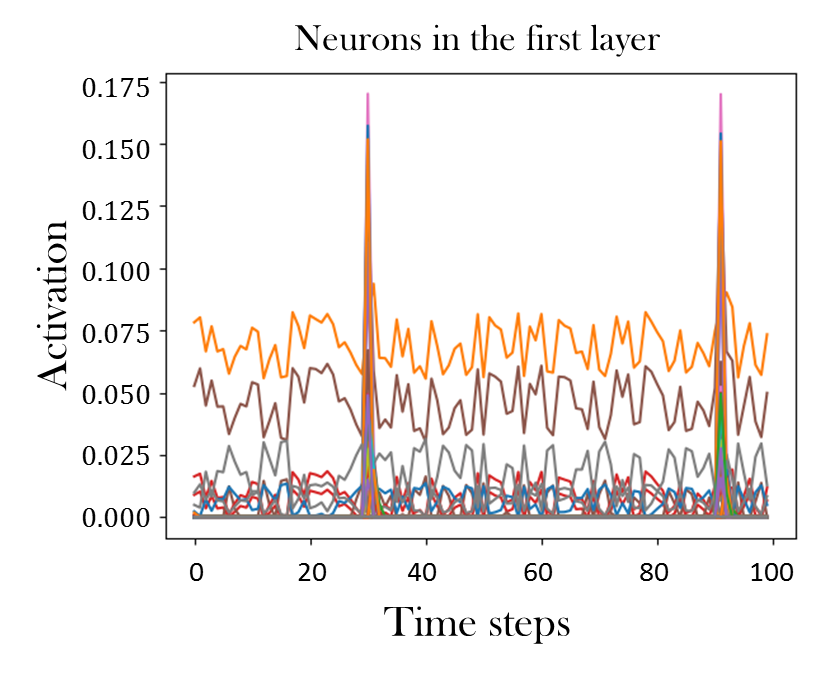}
			\caption{}
		\end{subfigure}
        \begin{subfigure}[b]{0.32\textwidth}
			\includegraphics[width=\textwidth]{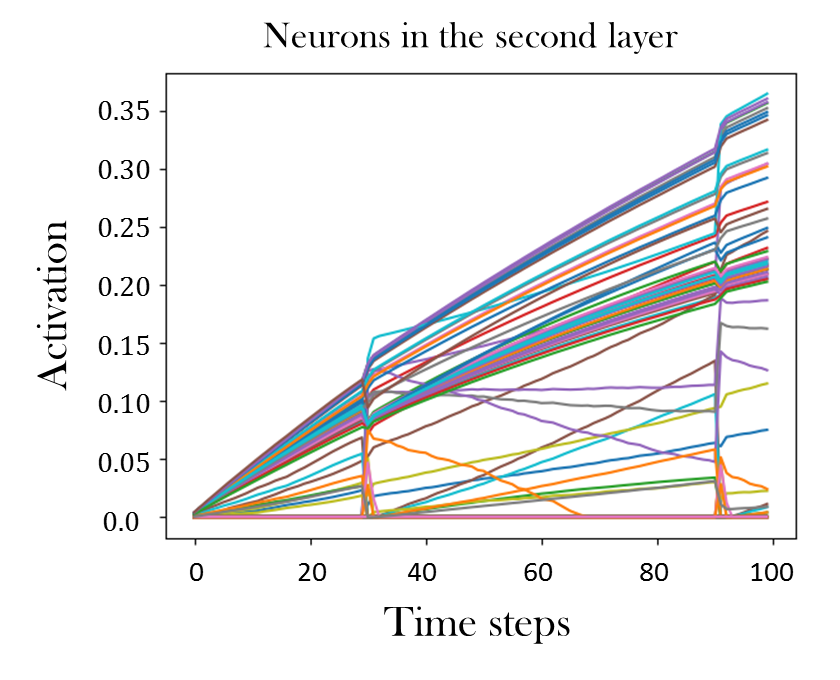}
			\caption{}
		\end{subfigure}
	\end{center}
	\caption{Neuron activations of DuRNN. The abscissa and ordinate are the same as  Fig. 4.}
\end{figure*}

Fig.4 and 5 show the neuron activations of IndRNN and DuRNN, respectively. A two-layer IndRNN is used as a simple naturally stacked multi-layer RNN to observe the neuron activations and the rules of dependence learning.
The results are shown in Fig. 4.
According to the theory in \cite{li2018independently}, if the recurrent weights in a IndRNN layer are  less than and stay at a distance from the one, this IndRNN layer can hardly capture the long-term dependence correspondingly.
So we reduce the upper bound of the constraints of the recurrent weight from $\sqrt[L]{2}$ (the setting of general case) to 0.9 on the first and second layers respectively in IndRNN for comparisons.
Using the 100-step adding problem, we can observe the neuron activations in the two layers visually and infer whether the dependence in sequence is learned by them.

Every subfigure in Fig. 4 shows the neuron activations in a layer with different constraints.
Respectively, (a), (b), (c) correspond to the first layer while (d), (e), (f) correspond to the second layer.
And (a), (d) correspond to the general 2-layer IndRNN; (b), (e) correspond to the IndRNN with reduced constraint on the first layer and (c), (f) correspond to the IndRNN with reduced constraint on the second layer.

As can be seen from Fig. 4, the general IndRNN can capture the sequential dependence in both of two layers.
When the constrain on the first layer is reduced, the long-term dependence learning capability of the second layer is not greatly affected, but the first layer can no longer capture long-term dependence when the constrain on the second layer is reduced.

Similar conclusions can be given when using other multi-layer RNN models.
We can conclude them as, in stacked multi-layer recurrent neural network, if some kind of dependence cannot be learned on some layer, the layers before it do not capture this dependence either.
In other words, if some layer has captured some kind of dependence, the following layers can also capture it.
That naturally leads to a corollary: long-term dependence tend to be learned on the layers which are close to the output, while the layers which are close to the input tend to preserving short-term dependence, and the dependence learning in a multi-layer RNN is progressive from the short to long.

Fig. 5 shows the neuron activations  in the first and second sublayers of DuRNN, which correspond to the full recurrent connections and independent recurrent connections, respectively.
The sequence length is also set as 100.
Compared with Figure 4(a-b), the curve in Fig. 5(a) is oscillating and the output changes suddenly, so the first sublayer of DuRNN only capture the short-term dependence. It should be noted that the numbers of active neurons in two layers are obviously increased.
Fig.5 illustrates that DuRNN can separate the dependence, which improves the utilization rate of neurons and make the feature representation more accurate.

\begin{figure*}
	\begin{center}
		\begin{subfigure}[b]{0.32\textwidth}
			\includegraphics[width=\textwidth]{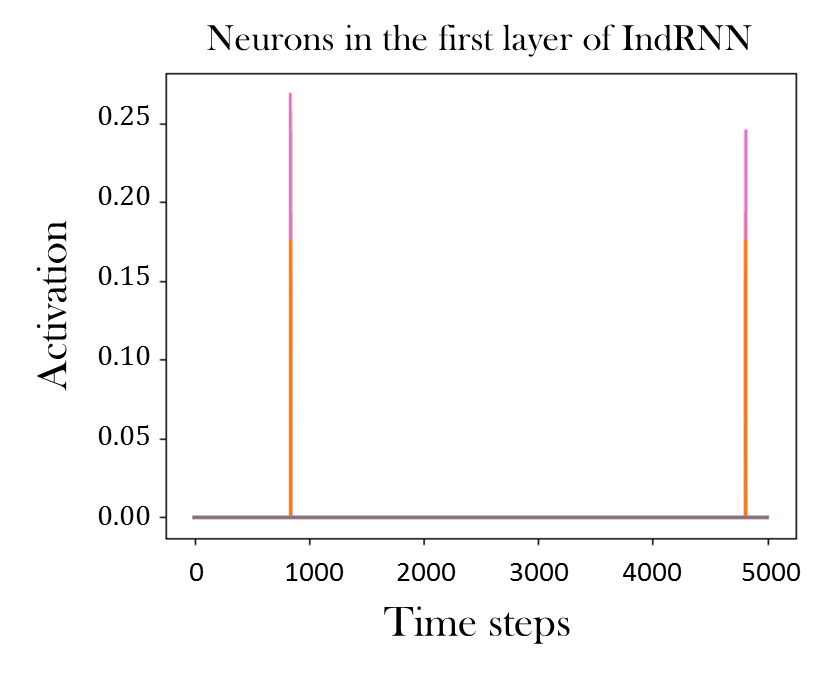}
			\caption{}
		\end{subfigure}
		\begin{subfigure}[b]{0.32\textwidth}
			\includegraphics[width=\textwidth]{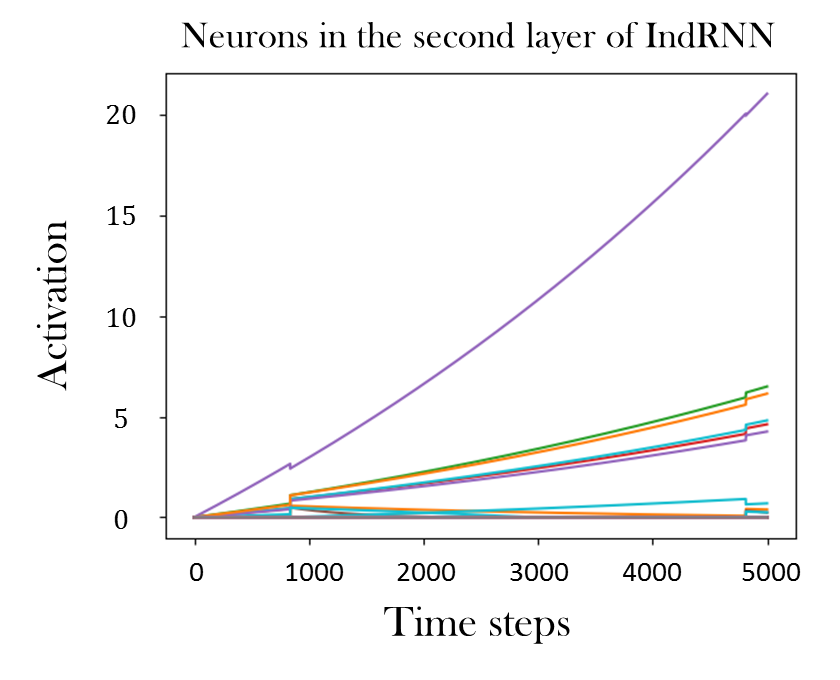}
			\caption{}
		\end{subfigure}
	\end{center}
	\begin{center}
		\begin{subfigure}[b]{0.32\textwidth}
			\includegraphics[width=\textwidth]{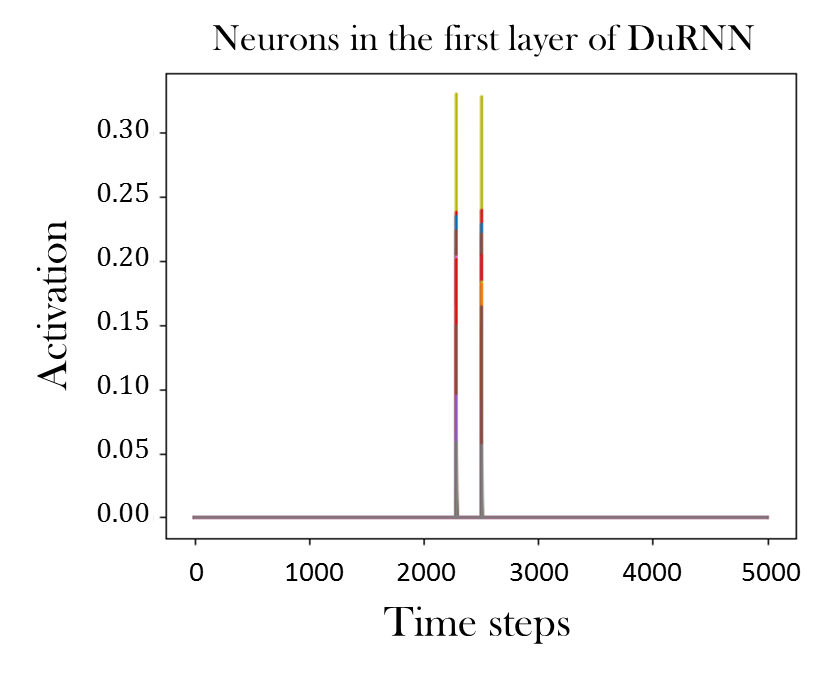}
			\caption{}
		\end{subfigure}
		\begin{subfigure}[b]{0.32\textwidth}
			\includegraphics[width=\textwidth]{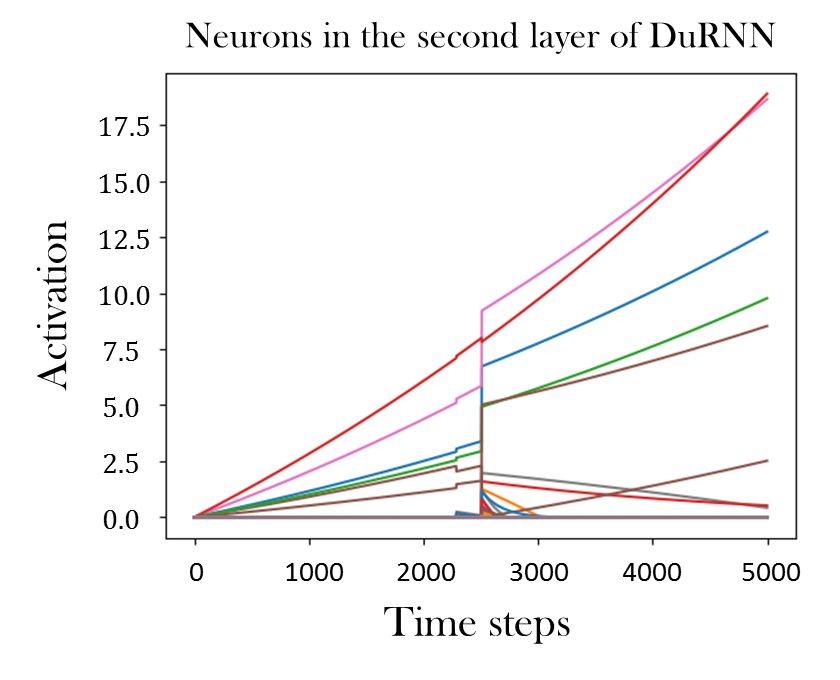}
			\caption{}
		\end{subfigure}
	\end{center}
	\caption{Neuron activations of IndRNN and DuRNN when dealing with very long sequences.}
\end{figure*}

Fig. 6 shows the neuron activations of the two networks on the 5000-step adding problem, which also have the above phenomena.
Our divide-and-conquer design is actually further strengthening the progressively learning characteristic of the multi-layer recurrent neural network and improves the efficiency and effect of learning various dependence.
Combined with the previous ablation analysis, it can be concluded that the divide-and-conquer strategy is reasonable and works very well.

\subsection{Other Tasks and Settings}

\textbf{Sequential MNIST classification} \cite{lecun1998gradient-based} is used to evaluate RNN models by changing the handwritten digits pixels to a sequence and using RNN to classify it.
The input of each RNN model is a sequence of length 784, which is obtained by expanding a $28 \times 28$ pixel matrix.
We can also randomly rearrange the sequence formed by each sample according to the same rule, make the problem more challenging, which is called permuted MNIST classification problem.

\textbf{Character-level Penn Treebank (PTB-c)} is to evaluate a RNN model by a language modeling task \cite{Treebank1993Building}.
This is a character-based sequence prediction task that is widely used to evaluate basic RNN dependence learning capability.
We set the environment parameter similar to \cite{Cooijmans2017Recurrent}.
The frame-wise batch normalization \cite{Laurent2016Batch} is applied to IndRNN and DuRNN.
Dropout \cite{Gal2016A} is used in the non-sequential direction.

\textbf{Skeleton based action recognition} is conducted with the skeleton data of the NTU RGB+D dataset \cite{Shahroudy_2016_NTURGBD}, which is one of the largest action recognition dataset.
This dataset contains 4 different modalities: Skeleton data, RGB videos, depth map sequences, 3D skeletal data and infrared videos.
The skeleton data contains 56881 sequences and 60 action classes,
and we evaluate our model in two standard protocols, i.e., Cross-Subject (CS) (40320 and 16560 samples for training and testing, respectively) and Cross-View (CV) (37920 and 18960 samples for training and testing, respectively).
In each evaluation protocol, $5\%$ of the training data are used as the validation data and from each skeleton action we sample the location information of skeleton joints in 20 frames as the input of the RNNs.

\textbf{Evaluation Metrics and Settings} are set as follows. Bit per character is to evaluate models in language modeling task, and cross entropy is used for evaluating others.
All tasks use Adam optimizer to train the model.
The learning rate is set to $2\times10^{-4}$ unless otherwise specified, which is to keep it consistent with all the tasks.
And it is the same with IndRNN so that the independent connection can be properly trained.
The parameters $\epsilon$ and $\gamma$ are set as 0.5 and 2, and $\epsilon$ is usually eased to 0 if it is not in the last layer of multi-layer models. The parameter $\delta$ is usually set as $\sqrt[L]{0.5}$.

\subsection{Comparison Results}

\subsubsection{MNIST classification}

In the sequential MNIST and permuted MNIST classification problems, the results of IRNN, uRNN \cite{arjovsky2016unitary}, RNN-path \cite{Neyshabur2016Path}, LSTM \cite{hochreiter1997long}, LSTM with recurrent dropout, LSTM with recurrent batch normalization and zoneout \cite{Semeniuta2016Recurrent,Cooijmans2017Recurrent,Krueger2017Zoneout} and IndRNN (6 layers) are listed for comparisons on this task.

\begin{table}\footnotesize
	\centering
	\caption{The error rates for different models on the sequential and permuted MNIST problems.}
	\begin{tabular*}{8.5cm}{lcc}
		\hline
		\qquad &\quad\qquad MNIST & PMNIST \\
		\hline
		IRNN  &\quad\qquad 5.0 & 18 \\
		uRNN  &\quad\qquad 4.9 & 8.6 \\
		RNN-path  &\quad\qquad 3.1 & - \\
		LSTM  &\quad\qquad 1.8 & 12 \\
		LSTM+Recurrent dropout  &\quad\qquad - & 7.5 \\
		LSTM+Recurrent batchnorm  &\quad\qquad 1.0 & 4.6 \\
		LSTM+Zoneout  &\quad\qquad - & 6.9 \\
		LSTM+Recurrent batchnorm+Zoneout  &\quad\qquad - & 4.1 \\
		IndRNN(6 layers)  &\quad\qquad 1.0 & 4.0 \\
		\hline
		\textbf{DuRNN(1 layer)}  &\quad\qquad 1.68 & 5.82 \\
		\textbf{DuRNN(3 layers)}  &\quad\qquad \textbf{0.96} & \textbf{3.46} \\
		\hline
	\end{tabular*}
\end{table}

In this task, our models are a 1-layer DuRNN and a 3-layer DuRNN. Since each DuRNN layer consists of two recurrent connections, a 3-layer DuRNN network corresponds to the 6-layer IndRNN. The contrast between these two models is fair and proper.
Furthermore, for our model, the batch size is set as 32, and 128 neurons are used in each recurrent connection.
Batch normalization (BN) \cite{ioffe2015batch} is used after the activation functions as an operation in both train and test task.
The learning rate is set as $2\times10^{-4}$. The results are shown in Table \uppercase\expandafter{\romannumeral1}.
DuRNN achieves the best performance at the same scale.
It can not only have a similar good effect in sequential experiments, but also further reduce the influence of random order on its dependence learning, which indicates that the dependence of various lengths can be better learned.

\subsubsection{Language Modeling}

\begin{table}
	\centering
	\caption{Character-level Penn Treebank}
	\begin{tabular*}{8.5cm}{lc}
		\hline
		\qquad  &\qquad\qquad\qquad\qquad Test Loss \\
		\hline
		RNN-tanh  &\qquad\qquad\qquad\qquad   1.55 \\
		RNN-relu  &\qquad\qquad\qquad\qquad   1.55 \\
		RNN-TRec  &\qquad\qquad\qquad\qquad   1.48 \\
		HF-MRNN  &\qquad\qquad\qquad\qquad   1.42 \\
		RNN-path  &\qquad\qquad\qquad\qquad   1.47 \\
		LSTM  &\qquad\qquad\qquad\qquad   1.36 \\
		LSTM+Recurrent dropout  &\qquad\qquad\qquad\qquad   1.32 \\
		LSTM+Recurrent batchnorm  &\qquad\qquad\qquad\qquad   1.32 \\
		HyperLSTM+LN  &\qquad\qquad\qquad\qquad   1.25 \\
		Hierarchical Multiscale LSTM+LN  &\qquad\qquad\qquad\qquad   1.24 \\
		LSTM+Zoneout  &\qquad\qquad\qquad\qquad   1.27 \\
		IndRNN(6 layers,50 steps) &\qquad\qquad\qquad\qquad   1.26 \\
        IndRNN(3 layers,150 steps) &\qquad\qquad\qquad\qquad   \textbf{1.23} \\
		\hline
        \textbf{DuRNN(1 layers, 50 steps)}  &\qquad\qquad\qquad\qquad   1.368\\
		\textbf{DuRNN(3 layers, 50 steps)}  &\qquad\qquad\qquad\qquad   1.247\\
		\textbf{DuRNN(3 layers, 100 steps)}  &\qquad\qquad\qquad\qquad   \textbf{1.230}\\
		\hline
	\end{tabular*}
\end{table}

In PTB-c task, DuRNN are compared with RNN-tanh \cite{krueger2016regularizing}, RNN-relu \cite{Neyshabur2016Path}, RNN-TRec \cite{krueger2016regularizing},
HF-MRNN \cite{Mikolov2012SUBWORD}, RNN-path, LSTM and LSTM with recurrent dropout, recurrent batchnorm, zoneout, HyperLSTM \cite{Ha2016HyperNetworks}, Hierarchical Multiscale LSTM \cite{chung2017hierarchical} and plain IndRNN with BN \cite{li2018independently}.
In this task we only consider the recurrent neural networks which are naturally stacked, since our paper focuses on the design of various sequential memory model instead of the application of RNN for a special task.

DuRNN of several layers are used for the task.
2000 hidden units are used for DuRNN, which is the same as IndRNN.
The dropout rate is set as 0.25 for neurons with independent connections and 0.3 for neurons with full recurrent connections over time.
The batch size is set as 128, if the training space is insufficient, it can be reduced.
BN is used before recurrent neurons to achieve better performance and turn out to make the model more robust than using it after the activation function.
The learning rate is $2\times10^{-4}$ and decayed by 5 after a proper delay when BPC on validation dataset no longer decreases.

The PTB-c task is measured by bit error. From Table \uppercase\expandafter{\romannumeral2}, we can find that DuRNNs obtain relatively good performance among these naturally stacked RNN models.
When the RNN with fewer layers is needed, DuRNN has a greater improvement on IndRNN, which can also be used as a lateral evidence that DuRNN is more efficient than IndRNN in assisting long-term dependence capture and expression.

\subsubsection{Action Recognition}

In skeleton based action recognition, RNN, LSTM, IndRNN and other skeleton based RNN methods:
JL\_d+RNN \cite{Zhang2017On}, S-trans+RNN, S-rota+RNN, S-trans+LSTM, VA-RNN \cite{zhang2019view}, PLSTM \cite{Shahroudy_2016_NTURGBD}, STA-LSTM, ST-LSTM+Trust Gate \cite{Liu2016Spatio} and pose conditioned STA-LSTM \cite{Baradel2017Pose} are compared with DuRNN.
In this task, we still focus on the comparisons of capacity of dependence capture between different sequential memory models, so we do not compare our model to the state-of-the-art methods which are specially designed using different kinds of neural networks.

In the model DuRNN, 512 neurons are used in each recurrent connections and the batch size is set as 128, which is the same as many other plain RNN models.
Sequence-wise normalization is used before both of the connections.
And normalization on the time axis is added after the independent recurrent connections to accelerate the training.
Dropout is applied with probability of 0.5 and 0.25 in the CS setting and 0.2 and 0.1 in the CV setting, respectively after each sublayer.
The learning rate is set as $2\times10^{-4}$ and decrease to 0.1 of the original when the classification accuracy  does not increase on the validation set.

\begin{table}\footnotesize
	\centering
	\caption{Skeleton based Action Recognition. The accuracies of plain RNN models on the CS and CV tasks are shown, and the multi-layer DuRNNs obtain good performance.}
	\begin{tabular*}{8.5cm}{lcc}
		\hline
		Method &\qquad\qquad\qquad CS & CV \\
		\hline

        RNN(1 layers)  &\qquad\qquad\qquad $56.02\%$ & $60.24\%$ \\
		RNN(2 layers)  &\qquad\qquad\qquad $56.29\%$ & $64.09\%$ \\
        JL\_d+RNN  &\qquad\qquad\qquad $70.26\%$ & $82.39\%$ \\
        S-trans+RNN  &\qquad\qquad\qquad $76.00\%$ & $82.30\%$ \\
        S-trans+RNN+aug  &\qquad\qquad\qquad $77.00\%$ & $85.00\%$ \\
        S-trans+S-rota+RNN  &\qquad\qquad\qquad $76.40\%$ & $85.40\%$ \\
        VA-RNN  &\qquad\qquad\qquad $79.40\%$ & $87.60\%$ \\
        VA-RNN+aug  &\qquad\qquad\qquad $79.80\%$ & $88.90\%$ \\
        \hline
		LSTM(1 layers)  &\qquad\qquad\qquad $59.84\%$ & $66.81\%$ \\
        LSTM(2 layers)  &\qquad\qquad\qquad $60.09\%$ & $67.29\%$ \\
		PLSTM(1 lyaers)  &\qquad\qquad\qquad $62.05\%$ & $69.40\%$ \\
        PLSTM(2 lyaers)  &\qquad\qquad\qquad $62.93\%$ & $70.27\%$ \\
		STA-LSTM  &\qquad\qquad\qquad $73.40\%$ & $81.20\%$ \\
        ST-LSTM+Trust Gate  &\qquad\qquad\qquad $69.20\%$ & $77.70\%$ \\
        S-trans+LSTM(6 layers)  &\qquad\qquad\qquad $76.60\%$ & $84.40\%$ \\
		Pose conditioned STA-LSTM  &\qquad\qquad\qquad $77.10\%$ & $84.50\%$ \\
        \hline
		IndRNN(4 layers)  &\qquad\qquad\qquad $78.58\%$ & $83.75\%$ \\
		IndRNN(6 layers)  &\qquad\qquad\qquad $81.80\%$ & $87.97\%$ \\
		\hline
        \textbf{DuRNN(1 layers)}  &\qquad\qquad\qquad $72.91\%$ & $77.53\%$ \\
        \textbf{DuRNN(2 layers)}  &\qquad\qquad\qquad $80.12\%$ & $85.50\%$ \\
		\textbf{DuRNN(3 layers)}  &\qquad\qquad\qquad $\textbf{82.53}\%$ & $\textbf{89.34}\%$ \\
        \hline
	\end{tabular*}
\end{table}

The test results are shown in Table \uppercase\expandafter{\romannumeral3}.
DuRNNs have the best performance on the test dataset.
In addition, we found that DuRNN with one layer can already work well.
It is very similar to LSTM, which also has a certain function of dependence partitioning.
It can be another evidence that processing dependence of different length separately is beneficial to save the resources.

\section{Conclusion}

In this paper we proposed a new sequential model named Dual Recurrent Neural Network (DuRNN).
It can capture various length dependence by a pair of different recurrent connections.  A selection mechanism is in charge of information transition between the short-term memory and long-term memory, which helps to improve the accuracy and the utilization of neurons.
The short-term memory is first learned and then selectively transferred to the long-term memory.
This mechanism functions like the macroscopical human memory in some sense, in which the short-term information are also selected and reinforced.
Its efficiency and rationality are validated and our model is applicable on both long- and short-term sequences.
Experimental results on multiple tasks show good performance of our model.



\ifCLASSOPTIONcaptionsoff
  \newpage
\fi

\bibliographystyle{IEEEtran}
\bibliography{R}

\begin{appendix}
\section{1}
The iterative algorithm used in gradient back propagation section is to better observe the relation between gradients and explain how ${\partial{h}_{t}}/{\partial{h}_{t-1}}$ drives the training of all variables.
We can get the following gradient formulas by bringing the variables adjacent to long-term memory into the original BPTT formula Eqs.(\ref{bptt_all}-\ref{btpp_t}),
\begin{equation*}
\begin{split}
\displaystyle\frac{\partial{Loss}}{\partial{U}}&=\sum_{t=0}^{L+1}\sum_{k=0}^{t+1}\frac{\partial{Loss_t}}{\partial{h_t}}\frac{\partial{h_t}}{\partial{h_k}}\frac{\partial{h_k}}{\partial{U}},\\
\displaystyle\frac{\partial{Loss}}{\partial{W_{s}}}&=\sum_{t=0}^{L+1}\sum_{k=0}^{t+1}\frac{\partial{Loss_t}}{\partial{h_t}}\frac{\partial{h_t}}{\partial{h_k}}\frac{\partial{h_k}}{\partial{Ws}},\\
\displaystyle\frac{\partial{Loss}}{\partial{W_{ss}}}&=\sum_{t=0}^{L+1}\sum_{k=0}^{t+1}\frac{\partial{Loss_t}}{\partial{h_t}}\frac{\partial{h_t}}{\partial{h_k}}\frac{\partial{h_k}}{\partial{S_k}}\frac{\partial{S_k}}{\partial{W_{ss}}},
\end{split}
\end{equation*}
\begin{equation*}
\begin{split}
\displaystyle\frac{\partial{Loss}}{\partial{W_{ls}}}&=\sum_{t=0}^{L+1}\sum_{k=0}^{t+1}\frac{\partial{Loss_t}}{\partial{h_t}}\frac{\partial{h_t}}{\partial{h_k}}\frac{\partial{h_k}}{\partial{S_k}}\frac{\partial{S_k}}{\partial{W_{ls}}},\\
\displaystyle\frac{\partial{Loss}}{\partial{b_{thre}}}&=\sum_{t=0}^{L+1}\sum_{k=0}^{t+1}\frac{\partial{Loss_t}}{\partial{h_t}}\frac{\partial{h_t}}{\partial{h_k}}\frac{\partial{h_k}}{\partial{S_k}}\frac{\partial{S_k}}{\partial{b_{thre}}}.
\end{split}
\end{equation*}

Furthermore, we also calculate the gradient of the variables adjacent to short-term memory according to Eq.(\ref{btpp_t}) based on the back propagation from both short-term and long-term memory,
\begin{equation*}
\begin{split}
\displaystyle\frac{\partial{Loss}}{\partial{W_{rec}}}&=\sum_{t=0}^{L+1}\sum_{k=0}^{t+1}\sum_{m=k}^{t+1}\frac{\partial{Loss_t}}{\partial{h_t}}\frac{\partial{h_t}}{\partial{h_m}}\frac{\partial{h_m}}{\partial{\widetilde{h}_{m}}}\frac{\partial{\widetilde{h}_{m}}}{\partial{\widetilde{h}_{k}}}\frac{\partial{\widetilde{h}_{k}}}{\partial{W_{rec}}},\\
\displaystyle\frac{\partial{Loss}}{\partial{W_{in}}}&=\sum_{t=0}^{L+1}\sum_{k=0}^{t+1}\sum_{m=k}^{t+1}\frac{\partial{Loss_t}}{\partial{h_t}}\frac{\partial{h_t}}{\partial{h_m}}\frac{\partial{h_m}}{\partial{\widetilde{h}_{m}}}\frac{\partial{\widetilde{h}_{m}}}{\partial{\widetilde{h}_{k}}}\frac{\partial{\widetilde{h}_{k}}}{\partial{W_{in}}}.\\
\end{split}
\end{equation*}
All of the above are direct results of bringing variables into the BPTT formula.
With the following specific gradient back propagation, the gradient of each variable can be directly calculated,
\begin{equation*}
\begin{split}
\displaystyle\frac{\partial{h_t}}{\partial{h_k}}&=\prod_{i=k+1}^{t}diag(U\circ \sigma_{s,i}^{'}),\\
\displaystyle\frac{\partial{h_k}}{\partial{S_k}}&=W_{s}^Tdiag(\widetilde{h}_{k}\circ \sigma_{s,k}^{'}),\\
\displaystyle\frac{\partial{h_t}}{\partial{U}}&=diag(h_{t-1}\circ \sigma_{s,t}^{'}),\\
\displaystyle\frac{\partial{h_t}}{\partial{W_s}}&=(\sigma_{s,t}^{'}\circ \vec{1})(S_t\circ \widetilde{h}_{t})^T,\\
\displaystyle\frac{\partial{S_k}}{\partial{W_{ls}}}&=(mm^{'}({W_{ss}}\widetilde{h}_{k}+{W_{ls}}h_{k-1}+b_{s})\\
\displaystyle&\quad \circ relu^{'}(mm({W_{ss}}\widetilde{h}_{k}+{W_{ls}}h_{k-1}+b_{s})\\
\displaystyle&\quad - b_{thre})\circ \vec{1})h_{k-1}^T,\\
\displaystyle\frac{\partial{S_k}}{\partial{W_{ss}}}&=(mm^{'}({W_{ss}}\widetilde{h}_{k}+{W_{ls}}h_{k-1}+b_{s})\\
\displaystyle&\quad \circ relu^{'}(mm({W_{ss}}\widetilde{h}_{k}+{W_{ls}}h_{k-1}+b_{s})\\
\displaystyle&\quad - b_{thre})\circ \vec{1})\widetilde{h}_{k}^T,\\
\displaystyle\frac{\partial{S_k}}{\partial{b_{thre}}}&=-\sum_{vector}relu^{'}(mm({W_{ss}}\widetilde{h}_{k}+{W_{ls}}h_{k-1}+b_{s})\\
\displaystyle&\quad-b_{thre}),\\
\displaystyle\frac{\partial{h_t}}{\partial{h_m}}&=\prod_{i=m+1}^{t}diag(U\circ \sigma_{s,i}^{'}),\\
\displaystyle\frac{\partial{h_m}}{\partial{\widetilde{h}_{m}}}&=diag(S_{m})W_{s}^Tdiag(\sigma_{s,m}^{'}),\\
\displaystyle\frac{\partial{\widetilde{h}_{m}}}{\partial{\widetilde{h}_{k}}}&=\prod_{i=k+1}^{m}C_\delta(W_{rec})^Tdiag(\sigma_{f,i}^{'}),\\
\end{split}
\end{equation*}
\begin{equation*}
\begin{split}
\displaystyle\frac{\partial{\widetilde{h}_{k}}}{\partial{W_{rec}}}&=(\sigma_{f,k}^{'}\circ \vec{1})\widetilde{h}_{k-1}^T,\\
\displaystyle\frac{\partial{\widetilde{h}_{k}}}{\partial{W_{in}}}&=(\sigma_{f,k}^{'}\circ \vec{1})x_{k}^T.
\end{split}
\end{equation*}
Compared with the iterative form, this gradient representation method can not reflect the relationship between the gradients, but it can show the composition of the gradient more clearly.

In addition, $\sum_{vector}$ is the same operation as in the gradient back propagation section, which can be expressed concretely as
\begin{align*}
\sum_{vector}v=\sum_{i=0}^Nv_i,
\end{align*}
where $v=(v_1,v_2,\cdots,v_N)^T\in\mathbb{R}^N$.
\end{appendix}

%
%
%

\end{document}